\pdfoutput=1

\documentclass[11pt]{article}

\usepackage{acl}

\usepackage{times}
\usepackage{latexsym}
\usepackage{bm}
\usepackage{amsmath}
\usepackage{amssymb}
\usepackage{algorithm}
\usepackage{algorithmic}
\usepackage{booktabs}
\usepackage{graphicx}
\usepackage{array}
\usepackage{multirow}

\usepackage[T1]{fontenc}

\usepackage[utf8]{inputenc}

\usepackage{microtype}

\title{A Simple Information-Based Approach to Unsupervised Domain-Adaptive Aspect-Based Sentiment Analysis}

\author{Xiang Chen, Xiaojun Wan \\
  Wangxuan Institute of Computer Technology, Peking University \\
  Center for Data Science, Peking University \\
  The MOE Key Laboratory of Computational Linguistics, Peking University \\
  \texttt{\{caspar,wanxiaojun\}@pku.edu.cn}}

\begin{document}
\maketitle
\begin{abstract}
Aspect-based sentiment analysis (ABSA) is a fine-grained sentiment analysis task which aims to extract the aspects from sentences and identify their corresponding sentiments. Aspect term extraction (ATE) is the crucial step for ABSA. Due to the expensive annotation for aspect terms, we often lack labeled target domain data for fine-tuning. To address this problem, many approaches have been proposed recently to transfer common knowledge in an unsupervised way, but such methods have too many modules and require expensive multi-stage preprocessing. In this paper, we propose a simple but effective technique based on mutual information maximization, which can serve as an additional component to enhance any kind of model for cross-domain ABSA and ATE. Furthermore, we provide some analysis of this approach. Experiment results\footnote{The code is available at \url{https://github.com/CasparSwift/DA_MIM}} show that our proposed method outperforms the state-of-the-art methods for cross-domain ABSA by 4.32\% Micro-F1 on average over 10 different domain pairs. Apart from that, our method can be extended to other sequence labeling tasks, such as named entity recognition (NER). 
\end{abstract}

\section{Introduction}
\label{introduction}

Aspect-Based Sentiment Analysis (ABSA)~\cite{liu2012sentiment, pontiki2015semeval} task can be split into two sub-tasks: Aspect Term Extraction (ATE) and Aspect Sentiment Classification (ASC). The former extracts the aspect terms from sentences while the latter aims to predict the sentiment polarity of every aspect term. ATE is considered to be a crucial step for ABSA because the errors of ATE may be propagated to the ASC task in the following stage. However, due to the expensive fine-grained token-level annotation for aspect terms, we often lack labeled training data for various domains, which becomes the major obstacle for ATE.

\begin{figure*}[t!]
    \centering
    \includegraphics[width=0.85\textwidth]{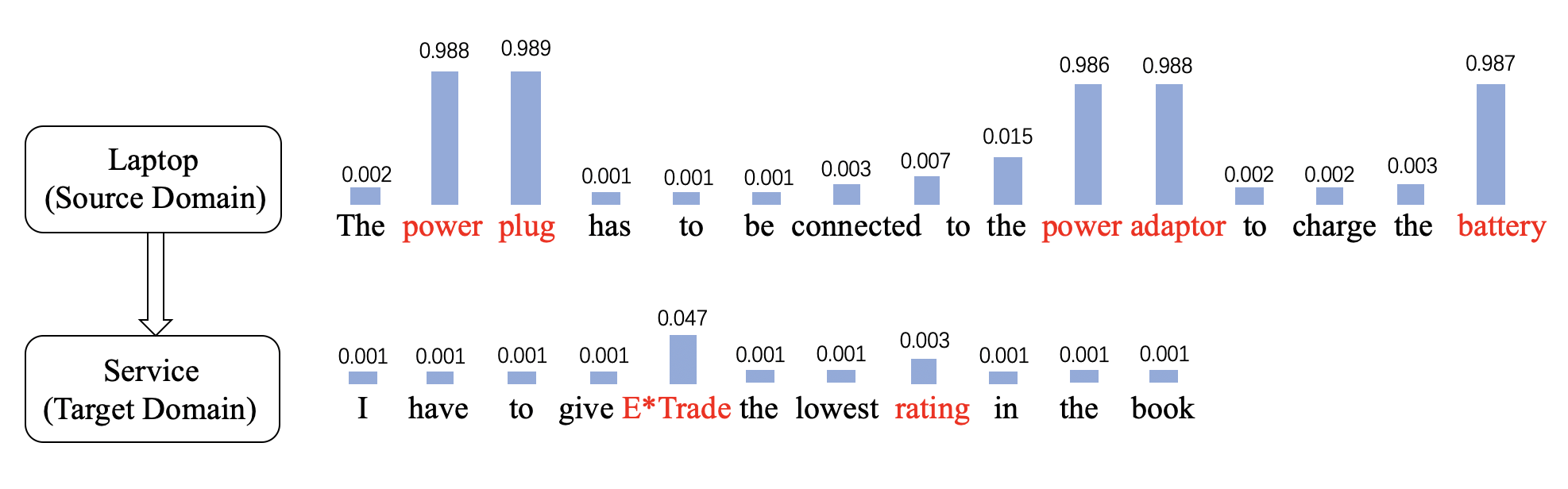}
    \caption{Examples of cross-domain ATE. The model is trained on the laptop domain. The value in this figure denotes the predicted probability of "a word belongs to an aspect" regardless of the sentiment polarity. The words in red indicate the ground truth aspect terms. The result shows that the model fails to extract any aspects in this sentence from service domain.}
    \label{LtoS}
\end{figure*}

To address such issue, previous studies follow the Unsupervised Domain Adaptation (UDA)~\cite{ramponi2020neural} scenario, which aims to transfer common knowledge from the source domain to the target domain. In UDA settings, we only have labeled source domain data and unlabeled target domain data. However, most aspect terms are strongly related to specific domains. The distribution of aspect terms may be significantly different across the domains, which causes performance degradation when transferring the domain knowledge. As shown in Figure~\ref{LtoS}, the model trained on the source domain (laptop) does not generalize well in the target domain (service). The model can easily extract the aspects related to laptop, such as "power plug", "power adaptor" and "battery", but it fails to extract the aspect terms "E*Trade" and "rating" that rarely appear in the laptop domain. Therefore, how to accurately discover the aspect terms from the unlabeled target domain data (raw texts) becomes the key challenge for cross-domain ABSA or ATE. 

Previous studies propose several approaches to tackle this problem. However, these methods still have some shortcomings in practical applications: (1) \textbf{Model Complexity}. Many existing approaches have multiple components, including domain classifier~\cite{li2019transferable,gong2020unified,chen2021bridge}, auto-encoder~\cite{wang2018recursive}, syntactically-aware self-attention~\cite{pereg2020syntactically}. Some studies introduce auxiliary tasks such as opinion co-extraction~\cite{ding2017recurrent,wang2018recursive,wang2019transferable,li2019transferable,pereg2020syntactically} and part-of-speech/dependency prediction~\cite{wang2018recursive,gong2020unified}.
Adding too many training objectives to the model may make it hard to optimize. Although these approaches are fancy and novel, we still need to seek for a simple but effective method according to the principle of Ockham's Razor. (2) \textbf{Multi-Stage Preprocessing}. Many previous methods require carefully designed multi-stage preprocessing, including non-lexical features extraction~\cite{jakob2010extracting,li2012cross,ding2017recurrent,wang2018recursive,pereg2020syntactically,gong2020unified,chen2021bridge} and target domain review generation~\cite{yu2021cross}. However, these preprocessing approaches are expensive when applied to real-world large scale datasets. Therefore, a single-stage method in an end-to-end manner is preferred. (3) \textbf{Extensibility}. All the above-mentioned methods are specifically designed for ABSA or ATE. However, essentially both ABSA and ATE can be formulated as sequence tagging tasks~\cite{mitchell2013open,zhang2015neural}. It's necessary to further investigate a unified technical scheme which can solve some other cross-domain extractive tagging tasks (e.g. named entity recognition (NER)).




In this paper, we get back to analyzing the intrinsic reason for the performance degradation when transferring aspect terms. From Figure~\ref{LtoS}, we have two important observations: (1) \textbf{Class Collapse}. The predictions tend to collapse into one single class (not an aspect term). (2) \textbf{Unconfident Predictions}. The predicted probabilities of ground truth aspects, namely "E*Trade" (0.047) and "rating" (0.003), are both slightly higher than other words. It seems that the model has the potential to identify correct aspects, but the prediction is not so confident.

Based on these two observations, in this paper, we propose a variant of the standard mutual information maximization technique~\cite{shi2012information,li2020rethinking,li2021cross} named "\textbf{FMIM}" (means \textbf{F}ine-grained \textbf{M}utual \textbf{I}nformation \textbf{M}aximization). The core idea is to maximize the token-level mutual information $I(X;\widehat{Y})=H(\widehat{Y})-H(\widehat{Y}|X)$, where $X$ denotes input tokens and $\widehat{Y}$ denotes their predicted labels. We maximize $H(\widehat{Y})$ to prevent the model from collapsing into one class, and minimize $H(\widehat{Y}|X)$ to enhance the confidence of model's predictions. Since it's difficult to precisely compute $H(\widehat{Y})$ because the joint distribution of $\widehat{Y}$ is intractable, FMIM uses a simple reduce mean approach to approximate it. FMIM is a general technique and can be added on top of any kinds of backbones or methods for cross-domain ABSA and ATE. Without adding any other modules or auxiliary tasks as the previous work did, all we do is to simply introduce an additional mutual information loss term, which achieves the model simplification and does not require any preprocessing. 

We find that FMIM is particularly effective for cross-domain ABSA and ATE.
The experiment results show that our method substantially exceeds the state-of-the-art~\cite{yu2021cross} by 4.32\% Micro-F1 (on average) over 10 domain pairs on ABSA task. Moreover, our method can be extended to other extractive tasks like cross-domain NER. We explore the effectiveness of our approach on cross-domain NER dataset and observe a considerable improvement over the state-of-the-art.


\section{Related Work}

\paragraph{Domain Adaptation} 
For sentiment analysis, existing domain adaptation methods mainly focus on coarse-grained sentiment classification: (1) Pivot-based methods~\cite{blitzer2006domain,pan2010cross,bollegala2012cross,yujiang2016learning,ziser2017neural,ziser-reichart-2018-pivot,ziser2019task} designed an auxiliary task of predicting pivots to transfer domain-invariant knowledge. (2) Adversarial methods~\cite{alam-etal-2018-domain,du2020adversarial} adopted Domain Adversarial Neural Network (DANN)~\cite{DANN}, which introduces a domain classifier to classify the domains of the instances. 
This method commonly serves as an important component of many state-of-the-art DA methods~\cite{du2020adversarial,chen2021bridge}. (3) Feature-based methods~\cite{fang2020cert,giorgi-etal-2021-declutr,li2021cross} introduced contrastive learning to learn domain-invariant features. For the sequence labeling task, we need token-level fine-grained features for the sentences. 

\paragraph{Cross-Domain ABSA}
Due to the accumulated errors between the two sub-tasks of ABSA (namely ATE and ASC), ABSA is typically combined together as a sequence labeling task~\cite{mitchell2013open,zhang2015neural,li2019unified}. Thus we need fine-grained domain adaptation for ABSA, which is more difficult than the coarse-grained one. \citet{jakob2010extracting} studied the cross-domain aspect extraction based on CRF. Another line of work \cite{li2012cross,ding2017recurrent,wang2018recursive,pereg2020syntactically,gong2020unified,chen2021bridge} utilized general syntactic or semantic relations to bridge the domain gaps, but they still relied on extra linguistic resources (e.g. POS tagger or dependency parser). \citet{li2019transferable} proposed a selective adversarial training method with a dual memory to align the words. However, adversarial training has been proven to be unstable~\cite{fedus2018many}.
FMIM considers cross-domain ABSA task from a brand-new perspective. We proved that only adding a mutual information maximization loss can substantially outperform all the above-mentioned methods with less complexity.

\paragraph{Mutual Information Maximization}
Mutual Information (MI) is a measure of the mutual dependency of two random variables in information theory~\cite{shannon1948mathematical}. Mutual Information Maximization (MIM) serves as a powerful technique for self-supervised learning~\cite{oord2018representation,hjelm2018learning,tschannen2019mutual} as well as semi-supervised learning~\cite{grandvalet2005semi}. Therefore, MIM can help to learn domain-invariant features for domain adaptation approaches. \citet{shi2012information} first proposed to maximize the MI between the target domain data and their estimated labels to learn discriminative clustering. Different from this approach, FMIM jointly optimizes the MI on both the source and target domains, which serves as an implicit alignment between the two domains. Moreover, most of the existing methods~\cite{shi2012information,khan2016adapting,li2020rethinking,li2021cross} only adopt MIM technique to deal with cross-domain image or sentiment classification tasks. To the best of our knowledge, this is the first work that illustrates the effectiveness of MIM for cross-domain sequence labeling tasks.

\section{Methodology}
\label{sec:methodology}
In this section, we first formulate our domain adaptation problem and introduce some notations. Then we present the proposed mutual information loss term and provide some analysis on it from both theoretical and empirical perspectives.

\subsection{Fine-Grained Mutual Information Maximization (FMIM)}
Let $\mathcal{D}_s$ and $\mathcal{D}_t$ denote the source domain training data and the target domain unlabeled data, respectively. For each sentence $X=\{x_1,...,x_n\}$, where $x_1,...,x_n$ denote the tokens, we have the predicted labels $\widehat{Y}=\{y_1,...,y_n\}$. Each $y_i$ is the label predicted by a model and $y_i\in \mathcal{S}$, where $\mathcal{S}=\{t_0,t_1,...,t_{T-1}\}$ is the tag set and $T=|\mathcal{S}|$. Specifically, for ABSA, the tag set is $\{$O, POS, NEU, NEG$\}$\footnote{Different from the previous work~\cite{li2019transferable,gong2020unified}, we adopt a different unified tagging scheme for ABSA instead of using $\{$B, I, O$\}$ to mark the aspect boundary. We extract the consecutive POS/NEU/NEG phrases as our final predictions. The experiment results show that the tagging scheme doesn't influence the performance.}, while for NER, the tag set is $\{$O, PER, ORG, LOC, MISC$\}$. 
Theoretically, the mutual information between a token $x$ and predicted label $y$ can be formulated as follows ($x$, $y$ are random variables here):
\begin{equation}
    \begin{aligned}
    I(x;y)&=H(y)-H(y|x)\\
    &=-\mathbb{E}_{y}[\log p(y)]+\mathbb{E}_{(x,y)}[\log p(y|x)]
    \end{aligned}
\label{mi}
\end{equation}

However, Eq~\ref{mi} is too complex to be precisely computed. We can use a mini-batch of data to approximate it. At each iteration of the training period, we randomly sample a mini-batch of data $\mathcal{B}_s$ from $\mathcal{D}_s$, and sample a mini-batch of data $\mathcal{B}_t$ from $\mathcal{D}_t$. Then, we collect and concatenate the model's outputs (the probability distributions over the tag set after softmax activation) of all samples from $\mathcal{B}_s$ and $\mathcal{B}_t$. After concatenation, we obtain an $N\times T$ tensor $M$, where $N$ equals to the sum of the token numbers of all samples. For illustration, we denote $X_{concat}=\{x_1,...,x_N\}$ as the concatenation of tokens of all samples. Then, the $(i,k)$-entry of the tensor $M$ indicates the conditional probability of the predicted label being tag $t_k$ given $i$-th token in $X_{concat}$, denoted as $M_{(i,k)}$.

For the first term of Eq~\ref{mi} (information entropy of $y$), we first calculate the distribution of the tags within the mini-batch $\mathcal{B}_s$ and $\mathcal{B}_t$. We define a tag probability $p(y=t_k)$ by the reduce-mean of the model outputs:
\begin{equation}
    p(y=t_k)\triangleq\frac{1}{N}\sum_{i=1}^N M_{(i,k)}
\end{equation}
Therefore, the first term can be approximated as:
\begin{equation}
    \Delta_1 = -\sum_{k=0}^{T-1}p(y=t_k)\log p(y=t_k)
\end{equation}

For the second term (negative conditional entropy), we can approximate it by the model's output probabilities as well:
\begin{equation}
    \Delta_2 = \frac{1}{N}\sum_{i=1}^N \sum_{k=0}^{T-1}M_{(i,k)}\log M_{(i,k)}
\end{equation}

Then we define our mutual information loss which is equivalent to the negative approximated mutual information. In practice, we do not expect $\Delta_1$ to be as large as possible. Thus, as suggested by~\citet{li2020rethinking,li2021cross}, we only maximize $\Delta_1$ when it is smaller than a pre-defined threshold $\rho$:
\begin{equation}
\mathcal{L}_{MI}=\begin{cases}
    -(\Delta_1+\Delta_2),\ \Delta_1<\rho\\
    -\Delta_2,\ \Delta_1\ge\rho
\end{cases}
\label{mi loss}
\end{equation}

The overall training objective is simply to jointly optimize the proposed MI loss $\mathcal{L}_{MI}$ and the original cross entropy loss $\mathcal{L}_{CE}$ for sequence labeling. We use a hyperparameter $\alpha$ to balance these two loss terms:
\begin{equation}
    \mathcal{L}_{train}=\mathcal{L}_{CE} + \alpha\mathcal{L}_{MI}
\label{train loss}
\end{equation}

\subsection{Analysis}
We can understand FMIM from the following three perspectives:

Firstly, by minimizing $\mathcal{L}_{MI}$ (i.e. maximizing $\Delta_1$ if $\Delta_1<\rho$), we keep $\Delta_1$ larger than a certain value $\rho$. We push the distribution of the predicted label $y$ (in mini-batches from both source and target) away from the 0-1 distribution $p(y=t_0)=1$ where $\Delta_1=0$. Consequently, we prevent the model from collapsing to a particular class and increase the diversity of the outputs~\cite{cui2020towards}. The model can extract more aspects in target domain, which can enhance the recall without reducing precision. Thus we solve the problem of the 
class collapse in section~\ref{introduction}.

Secondly, by minimizing $\mathcal{L}_{MI}$ (i.e. maximizing $\Delta_2$, namely, minimizing the conditional entropy). We encourage the model to make more confident predictions. Thus we solve the problem of unconfident predictions in section~\ref{introduction}. Moreover, minimizing the conditional entropy intuitively enlarges the margin between different classes, which makes the decision boundary learned on source domain data easier to fall into the margin~\cite{grandvalet2005semi,li2020rethinking,li2021cross}. This is beneficial to the domain transferring.

Thirdly, MIM is a commonly used technique in unsupervised learning or self-supervised learning (SSL)~\cite{hjelm2018learning,tschannen2019mutual}. According to the results given by~\citet{oord2018representation}, mutual information is an upper bound of negative InfoNCE which is a loss function widely used in contrastive learning~\cite{he2020momentum,chen2020simple,chen2020simCLRv2,chen2020mocov2}:
\begin{equation}
    I(X;Y)\ge C-\mathcal{L}_{NCE}(X,Y)
\end{equation}
where $C$ is a constant. Therefore minimizing the InfoNCE is equivalent to maximizing mutual information. In other words, introducing $\mathcal{L}_{MI}$ can be viewed as an implicit way of contrastive learning. 

\section{Experiments}
To evaluate the effectiveness of FMIM technique introduced in Section~\ref{sec:methodology}, we apply our method to cross-domain ABSA, ATE and NER datasets.

\subsection{Experiment Setup}
\paragraph{Datasets} Our experiment is conducted on four benchmarks with different domains: Laptop ($\mathbb{L}$), Restaurant ($\mathbb{R}$), Device ($\mathbb{D}$), and Service ($\mathbb{S}$). $\mathbb{L}$ and $\mathbb{R}$ are from SemEval ABSA challenge~\cite{pontiki-etal-2014-semeval,pontiki2015semeval,pontiki2016semeval}. $\mathbb{D}$ is provided
by~\citet{hu2004mining} and contains digital product reviews. $\mathbb{S}$ is provided by~\citet{toprak2010sentence} and contains reviews from web services.

It's worth noting that there are two different dataset settings in previous studies, so we evaluate our method on both of them. For ABSA task, previous work~\cite{li2019unified,li2019transferable,gong2020unified,yu2021cross} conducted experiments for 10 domain pairs on the above-mentioned four domains. For ATE task only, previous work~\cite{wang2018recursive,wang2019transferable,pereg2020syntactically,chen2021bridge} conducted experiments for 6 domain pairs on $\mathbb{L}$, $\mathbb{R}$ and $\mathbb{D}$. They use three different data splits with a fixed train-test ratio 3:1. Apart from that, the amount of sentences of some domains are different. Detailed statistics are shown in Table~\ref{table: statistics1}  and~\ref{table: statistics2}.

\begin{table}[t!]
\centering
\small
\begin{tabular}{l|c|c|c}
\hline
\bf Domain & \bf Sentences & \bf Train & \bf Test \\
\hline
Laptop ($\mathbb{L}$) & 3845 & 3045 & 800 \\
\hline
Restaurant ($\mathbb{R}$) & 6035 & 3877 & 2158 \\
\hline
Device ($\mathbb{D}$) & 3836 & 2557 & 1279 \\
\hline
Service ($\mathbb{S}$) & 2239 & 1492 & 747 \\
\hline
\end{tabular}
\caption{\label{table: statistics1}Statistics of our cross-domain ABSA datasets.}
\end{table}

\begin{table}[t!]
\centering
\small
\begin{tabular}{l|c|c|c}
\hline
\bf Domain & \bf Sentences & \bf Train & \bf Test \\
\hline
Laptop ($\mathbb{L}$) & 3845 & 2884 & 961 \\
\hline
Restaurant ($\mathbb{R}$) & 5841 & 4381 & 1460 \\
\hline
Device ($\mathbb{D}$) & 3836 & 2877 & 959 \\
\hline
\end{tabular}
\caption{\label{table: statistics2}Statistics of our cross-domain ATE datasets.}
\end{table}

\begin{table}[t!]
\centering
\small
\begin{tabular}{l|c|c}
\hline
\bf Domain & CoNLL2003 & CBS News \\
\hline
\bf Train (labeled) & 15.0K & - \\
\hline
\bf Train (unlabeled) & - & 398,990\\
\hline
\bf Dev & 3.5K & - \\
\hline
\bf Test & 3.7K & 2.0K\\
\hline
\end{tabular}
\caption{\label{table: ner statistics}Statistics of our cross-domain NER datasets.}
\end{table}

For cross-domain NER, following the same dataset setting of~\citet{jia2019cross,jia2020multi}, we take CoNLL2003 English dataset~\cite{sang2003introduction} and CBS SciTech News dataset collected by~\citet{jia2019cross} as the source and target domain data, respectively. Detailed statistics of the datasets are shown in Table~\ref{table: ner statistics}. 

\paragraph{Evaluation} For ABSA, all the experiments are repeated 5 times with 5 different random seeds and we report the Micro-F1 over 5 runs, which is the same as the previous work. Only correct aspect terms with correct sentiment predictions can be considered to be true positive instances. For ATE, following~\citet{chen2021bridge}, we report the mean F1-scores of aspect terms over three splits with three random seeds (9 runs for each domain pair). For NER, we report the F1-score of named entities.

\paragraph{Implementation Details}
For all the tasks, we use the pre-trained BERT-base-uncased~\cite{devlin2018bert} model provided by HuggingFace~\cite{wolf2019huggingface} as our feature extractor. The maximum input length of BERT is 128. Our sentiment classifier is a MLP with two hidden layers with hidden size 384. We take ReLU as the activation function. For the optimization of model parameters, we use the AdamW~\cite{loshchilov2018fixing} as the optimizer with a fixed learning rate of $2e-5$ or $1e-5$. We train the model for 20 epochs for ABSA and 3 epochs for NER.

For ABSA, we set $\alpha=0.005,\rho=0.5$ for $\mathbb{R}\to\mathbb{D}$, $\alpha=0.01,\rho=0.25$ for $\mathbb{S}\to\mathbb{L}$, $\alpha=0.015,\rho=0.7$ for $\mathbb{R}\to\mathbb{L}$, $\alpha=0.025,\rho=0.5$ for $\mathbb{L}\to\mathbb{S}$ and $\alpha=0.01,\rho=0.5$ for the rest of domain pairs. We set $\alpha=0.009,\rho=0.5$ for cross-domain NER\footnote{Our results can be improved by tuning the hyperparameters carefully, but this is not what we mainly focus on.}.

For ATE, the hyperparameter settings are presented in Table~\ref{hyper}.

\begin{table}[t!]
    \centering
    \resizebox{0.48\textwidth}{!}{
    \begin{tabular}{l|c|c|c|c|c|c}
    \toprule[1.5pt]
        \bf Hyperparameter & \bf R$\to$L & \bf L$\to$R & \bf R$\to$D & \bf D$\to$R & \bf L$\to$D & \bf D$\to$L \\
        \midrule[1pt]
        $\alpha$ & 0.015 & 0.01 & 0.015 & 0.01 & 0.015 & 0.01  \\
        $\rho$ & 0.7 & 0.5 & 0.2 & 0.5 & 0.2 & 0.5 \\
        weight decay & 0.1 & 0.1 & 1 & 0.1 & 1 & 0.1 \\
        batch size & 16 & 16 & 16 & 16 & 16 & 16 \\ 
        \bottomrule[1.5pt]
    \end{tabular}
    }
    \caption{Hyperparameter settings for ATE.}
    \label{hyper}
\end{table}

\begin{table*}[t!]
\centering
\resizebox{0.98\textwidth}{!}{
\begin{tabular}{l|ccc|ccc|cc|cc|c}
\toprule[1.5pt]
\bf Methods & \bf S$\to$R & \bf L$\to$R & \bf D$\to$R & \bf R$\to$S & \bf L$\to$S & \bf D$\to$S & \bf R$\to$L & \bf S$\to$L & \bf R$\to$D & \bf S$\to$D & \bf Avg. \\
\midrule[1pt]
Hier-Joint~\cite{ding2017recurrent} & 31.10 & 33.54 & 32.87 & 15.56 & 13.90 & 19.04 & 20.72 & 22.65 & 24.53 & 23.24 & 23.71 \\
RNSCN~\cite{wang2018recursive} & 33.21 & 35.65 & 34.60 & 20.04 & 16.59 & 20.03 & 26.63 & 18.87 & 33.26 & 22.00 & 26.09 \\
AD-SAL~\cite{li2019transferable} & 41.03 & 43.04 & 41.01 & 28.01 & 27.20 & 26.62 & 34.13 & 27.04 & 35.44 & 33.56 & 33.71 \\
\midrule[1pt]
BERT-Base$^*$~\cite{devlin2018bert} & 44.76 & 26.88 & 36.08 & 19.41 & 27.27 & 27.62 & 28.95 & 29.20 & 29.47 & 33.96 & 30.36 \\
BERT-Base~\cite{gong2020unified} & 44.66 & 40.38 & 40.32 & 19.48 & 25.78 & 30.31 & 31.44 & 30.47 & 27.55 & 33.96 & 32.43 \\
BERT-DANN~\cite{gong2020unified} & 45.84 & 41.73 & 34.68 & 21.60 & 25.10 & 18.62 & 30.41 & 31.92 & 34.41 & 23.97 & 30.83 \\
BERT-UDA~\cite{gong2020unified} & 47.09 & 45.46 & 42.68 & 33.12 & 27.89 & 28.03 & 33.68 & \textbf{34.77} & 34.93 & 32.10 & 35.98 \\
CDRG (Indep)~\cite{yu2021cross} & 44.46 & 44.96 & 39.42 & 34.10 & 33.97 & 31.08 & 33.59 & 26.81 & 25.25 & 29.06 & 34.27 \\
CDRG (Merge)~\cite{yu2021cross} & 47.92 & 49.79 & 47.64 & 35.14 & 38.14 & 37.22 & \bf 38.68 & 33.69 & 27.46 & 34.08 & 38.98 \\

BERT-Base + FMIM (ours) & \textbf{50.20} & \textbf{53.24} & \textbf{54.98} & \textbf{42.78} & \textbf{43.20} & \textbf{46.69} & 38.20 & 32.49 & \textbf{35.87} & \textbf{35.38} & \bf 43.30$^\dagger$ \\
\bottomrule[1.5pt]
\end{tabular}
}
\caption{The results for cross-domain ABSA task\footnote{Following~\citet{gong2020unified}, we do not remove the samples which do not contain any explicit aspects for fair comparison.}. The evaluation metric is based on Micro-F1. BERT-base$^*$ is our implementation by using a vanilla BERT. $\dagger$ indicates that our result significantly outperforms CDRG (Merge) based on t-test ($p<0.01$).}
\label{tab:absa results}
\end{table*}

\begin{table*}[t!]
\centering
\resizebox{0.98\textwidth}{!}{
\begin{tabular}{l|ccc|ccc|cc|cc|c}
\toprule[1.5pt]
\bf Methods & \bf S$\to$R & \bf L$\to$R & \bf D$\to$R & \bf R$\to$S & \bf L$\to$S & \bf D$\to$S & \bf R$\to$L & \bf S$\to$L & \bf R$\to$D & \bf S$\to$D & \bf AVG \\
\midrule[1pt]
Hier-Joint~\cite{ding2017recurrent} & 46.39 & 48.61 & 42.96 & 27.18 & 25.22 & 29.28 & 34.11 & 33.02 & 34.81 & 35.00 & 35.66 \\
RNSCN~\cite{wang2018recursive} & 48.89 & 52.19 & 50.39 & 30.41 & 31.21 & 35.50 & 47.23 & 34.03 & \textbf{46.16} & 32.41 & 40.84 \\
AD-SAL~\cite{li2019transferable} & 52.05 & 56.12 & 51.55 & 39.02 & 38.26 & 36.11 & 45.01 & 35.99 & 43.76 & \textbf{41.21} & 43.91 \\
\midrule[1pt]
BERT-Base$^*$~\cite{devlin2018bert} & 54.93 & 30.98 & 40.15 & 22.92 & 31.63 & 31.27 & 35.07 & 36.96 & 32.08 & 38.17 & 35.42 \\
BERT-Base~\cite{gong2020unified} & 54.29 & 46.74 & 44.63 & 22.31 & 30.66 & 33.33 & 37.02 & 36.88 & 32.03 & 38.06 & 37.59 \\
BERT-DANN~\cite{gong2020unified} & 54.32 & 48.34 & 44.63 & 25.45 & 29.83 & 26.53 & 36.79 & 39.89 & 33.88 & 38.06 & 37.77 \\
BERT-UDA~\cite{gong2020unified} & 56.08 & 51.91 & 50.54 & 34.62 & 32.49 & 34.52 & 46.87 & \textbf{43.98} & 40.34 & 38.36 & 42.97 \\
CDRG (Indep)~\cite{yu2021cross} & 53.79 & 55.13 & 50.07 & 41.74 & 44.14 & 37.10 & 40.18 & 33.22 & 30.78 & 34.97 & 42.11 \\
CDRG (Merge)~\cite{yu2021cross} & 56.26 & 60.03 & 52.71 & 42.36 & 47.08 & 41.85 & 46.65 & 39.51 & 32.60 & 36.97 & 45.60 \\
BERT-Base + FMIM (ours) & \textbf{59.24} & \textbf{63.41} & \textbf{57.29} & \textbf{51.35} & \textbf{54.92} & \textbf{52.85} & \textbf{49.42} & 42.44 & 39.72 & 37.62 & \bf 50.83$^\dagger$ \\
\bottomrule[1.5pt]
\end{tabular}
}
\caption{The results for the sub-task of ATE based on Micro-F1. $\dagger$ indicates that our result significantly outperforms CDRG (Merge) based on t-test ($p<0.01$).}
\label{tab:ae results}
\end{table*}

\subsection{Baselines \& Compared Methods}

\paragraph{Cross-Domain ABSA} Hier-Joint~\cite{ding2017recurrent} use manually designed syntactic rule-based auxiliary tasks. RNSCN~\cite{wang2018recursive} is based on a novel recursive neural structural correspondence network. And an auxiliary task is designed to predict the dependency relation between any two adjacent words. AD-SAL~\cite{li2019transferable} dynamically learn an alignment between words by adversarial training. BERT-UDA~\cite{gong2020unified} incorporates masked POS prediction, dependency relation prediction and instance reweighting. BERT-Base~\cite{devlin2018bert} indicates directly fine-tuning BERT-base-uncased model on the source training data.
BERT-DANN~\cite{gong2020unified} performs adversarial training on each word in the same way as~\citet{DANN}. CDRG~\cite{yu2021cross} generates the target domain reviews with independent and merge training strategies.

\paragraph{Cross-Domain ATE} BERT-UDA can be modified for ATE by remapping the B/I/O labels. SA-EXAL~\cite{pereg2020syntactically} incorporates syntactic information with attention mechanism. BERT-Cross~\cite{xu-etal-2019-bert} conducts BERT post-training on a combination of Yelp and Amazon corpus. BaseTagger~\cite{chen2021bridge} is a strong baseline which takes CNN as its backbone. SemBridge~\cite{chen2021bridge} uses semantic relations to bridge the domain gap.

\paragraph{Cross-Domain NER} Cross-Domain LM~\cite{jia2019cross} designs parameter generation network and performs cross-domain language modeling. Multi-Cell LSTM~\cite{jia2020multi} designs a multi-cell LSTM structure to model each entity type using a separate cell state.

\begin{table*}[t!]
\centering
\resizebox{0.98\textwidth}{!}{
\begin{tabular}{l|c|c|c|c|c|c|c|c}
\toprule[1.5pt]
\bf Methods & \bf Embedding & \bf R$\to$L & \bf L$\to$R & \bf R$\to$D & \bf D$\to$R & \bf L$\to$D & \bf D$\to$L & \bf Avg. \\
\midrule[1pt]
BERT-UDA~\cite{gong2020unified} & BERT$_{B}$ & 44.24 & 50.52 & 40.04 & 53.39 & 41.48 & 52.33 & 47.00 \\
SA-EXAL~\cite{pereg2020syntactically} & BERT$_{B}$ & 47.59 & 54.67 & 40.50 & 54.54 & 42.19 & 47.72 & 47.87 \\
BERT-Cross~\cite{xu-etal-2019-bert} & BERT$_{E}$ & 46.30 & 51.60 & \bf 43.68 & 53.15 & 44.22 & 50.04 & 48.17 \\
\midrule[1pt]
BaseTagger~\cite{chen2021bridge} & Word2vec & 48.86 & 61.42 & 40.56 & 57.67 & 43.75 & 51.95 & 50.70 \\
BaseTagger + FMIM & Word2vec & 49.74 & 65.60 & 40.64 & 59.38 & 44.22 & 51.87 & 51.91 \\
\midrule[1pt]
SemBridge~\cite{chen2021bridge} & Word2vec & 51.53 & 65.96 & 43.03 & 60.61 & 45.37 & \bf 53.77 & 53.38 \\
SemBridge + FMIM & Word2vec & 49.00 & 67.41 & 43.10 & 63.30 & \bf 45.68 & 53.00 & \bf 53.58 \\
\midrule[1pt]
BERT-Base~\cite{devlin2018bert} & BERT$_{B}$ & 33.89 & 42.74 & 35.30 & 36.86 & 43.54 & 46.06 & 39.73 \\
BERT-Base + FMIM & BERT$_{B}$ & \bf 52.00 & \bf 71.63 & 38.73 & \bf 65.18 & 44.62 & 49.46 & \bf 53.58 \\
\bottomrule[1.5pt]
\end{tabular}
}
\caption{The results for cross-domain ATE task. BERT$_B$ indicates BERT-Base model and BERT$_E$ is post-trained by~\citet{xu-etal-2019-bert}. The metric is mean F1-score over 9 runs for each domain pair.}
\label{tab:ae new results}
\end{table*}

\begin{table*}[t!]
    \centering
    \small
    \begin{tabular}{l|c|c}
    \toprule[1.5pt]
    \bf Methods & \bf Micro-F1 & \bf Raw Texts of Target Domain \\
    \midrule[1pt]
    Cross-Domain LM~\cite{jia2019cross} & 73.59 & 18,474K \\
    Multi-Cell LSTM~\cite{jia2020multi} & 72.81 & 1.931K \\
    Multi-Cell LSTM (All)~\cite{jia2020multi} & 73.56 & 8,664K \\
    BERT-Base~\cite{devlin2018bert} & 74.23 & - \\
    BERT-Base + FMIM (ours) & \bf 75.32 & $\le$45K \\
    \bottomrule[1.5pt]
    \end{tabular}
    \caption{The results for cross-domain NER task based on Micro-F1. "Multi-Cell LSTM (All)" indicates using full set of the target domain raw texts for language modeling~\cite{jia2020multi}. Despite that we have 399K target domain raw text as shown in Table~\ref{table: ner statistics}, there are only no more than 45K of them will be fed into the model. The reason is that we only randomly sample a part of raw texts (15K, the same amount as source training data) at each epoch and we train for only 3 epochs.}
    \label{tab: ner result}
\end{table*}

\subsection{Results for Cross-Domain ABSA}
The overall results for cross-domain ABSA are shown in Table~\ref{tab:absa results}. As the previous work did, we conduct our experiments on 10 different domain pairs. 
We observe that BERT-Base+FMIM outperforms the state-of-the-art method CDRG~\cite{yu2021cross} in most of domain pairs except when $\mathbb{L}$ is the target domain. Our approach achieve 1.3\%$\sim$9.47\% absolute improvement of Micro-F1 compared to CDRG (Merge). When taking $\mathbb{S}$ as the target domain, we can obtain 7.64\%, 5.06\% and 9.47\% improvement respectively. 

Following~\citet{gong2020unified,yu2021cross}, we also provide the results for the ATE sub-task in Table~\ref{tab:ae results}. We can observe that FMIM can consistently improve the performance of ATE on most of the domain pairs and achieves an average improvement of 5.23\% Micro-F1 compared to CDRG.

Furthermore, from the results in Tables~\ref{tab:absa results} and~\ref{tab:ae results}, we have the following observations:

(1) The vanilla BERT-base model~\cite{devlin2018bert} can beat the previous models based on RNN or LSTM (Hier-Joint~\cite{ding2017recurrent}, RNSCN~\cite{wang2018recursive}) and it has a competitive performance with AD-SAL~\cite{li2019transferable}, which shows that the language model pre-trained on large-scale corpora has the generalization ability across domains to some extent. But this result still can be improved by some specific domain adaptation techniques.

(2) The improvement of BERT-DANN is quite marginal and inconsistent across 10 domain pairs compared to the vanilla BERT-base model. This is reasonable because BERT-DANN discriminates the domains in word level, which cannot capture the semantic relations between words. Moreover, many common words may appear in both source and target domain, and classifying the domains of these words will unavoidably introduce noise to the model and result in unstable training.

(3) FMIM substantially outperforms CDRG~\cite{yu2021cross}, the state-of-the-art method for cross-domain ABSA. 
We think the reason for this improvement is that simply generating target domain review data may not directly address the class collapse and unconfident predictions problems. However, FMIM is entirely orthogonal with CDRG, so adding it on the top of CDRG can possibly achieve better performance. 

\subsection{Results for Cross-Domain ATE}
Table~\ref{tab:ae new results} shows the results for cross-domain ATE. In this section, we illustrate the effectiveness of adding FMIM to other methods. All the methods with FMIM outperforms the BERT-UDA, SA-EXAL and BERT-Cross baselines. When adding on top of BaseTagger, we can observe an absolute improvement of 1.21\%. For the state-of-the-art SemBridge method, we improve the F1-score by 1.45\%, 2.69\%, 0.31\% on $\mathbb{L}\to\mathbb{R}$, $\mathbb{D}\to\mathbb{R}$, $\mathbb{L}\to\mathbb{D}$. When taking the vanilla BERT-base as our backbone, FMIM can achieve an improvement of 13.85\%. This results illustrates that FMIM can serve as an effective technique to enhance common cross-domain ATE models.

\subsection{Results for Cross-Domain NER}
The experiment results on unsupervised domain adaptation of NER are presented in Table~\ref{tab: ner result}. Due to the similarity of NER task and aspect term extraction task, FMIM based on BERT-Base~\cite{devlin2018bert} can still outperform the state-of-the-art cross-domain NER model Multi-Cell LSTM~\cite{jia2020multi} by 1.76\% F1-scores. FMIM also exceeds the baseline of directly using BERT-Base by 1.09\% F1-scores. It's worth noting that the amount the raw texts of target domain we used is 192 times less than that of~\citet{jia2020multi}, which shows the great data efficiency of FMIM.


\begin{table*}[t!]
    \centering
    \resizebox{0.98\textwidth}{!}{
    \begin{tabular}{c|ccc|c}
    \toprule[1.5pt]
    \bf Method & $H(\widehat{Y})$ & $H(\widehat{Y}|X)$ & $I(X;\widehat{Y})$ & \bf Predictions \\
    \midrule[1pt]
    \multicolumn{5}{l}{Sentence: \textit{\textbf{Trading} through \textbf{e*trade} is fairly easy.}} \\
    \hline
    BERT-Base  & 0.54 & 0.33 & 0.21 & None$_\times$ \\
    \hline
    BERT-Base + FMIM (ours) & 0.88 & 0.04 & 0.84 & Trading (POS)$_\checkmark$, e*trade (POS)$_\checkmark$ \\
    \midrule[1pt]
    \multicolumn{5}{l}{Sentence: \textit{The few problems that I have had with \textbf{Etrade} are mainly concerning delayed \textbf{trade confirmations}.}}\\
    \hline
    BERT-Base & 0.22 & 0.13 & 0.09 & None$_\times$ \\
    \hline
    BERT-Base + FMIM (ours) & 0.74 & 0.06 & 0.68 & Etrade (NEG)$_\checkmark$, trade confirmations (NEG)$_\checkmark$ \\
    \bottomrule[1.5pt]
    \end{tabular}
    }
    \caption{Two examples from the test set of service domain in $\mathbb{L}\to\mathbb{S}$ settings. The words in bold are the ground truth aspect terms. "POS" and "NEG" are the sentiment predictions.}
    \label{tab: case study}
\end{table*}

\begin{table*}[t!]
    \centering
    \begin{tabular}{|l|}
    \hline
    E1: facilities including a comprehensive [\textbf{glossary}]$_\times$  \textbf{of} [\textbf{terms}]$_\times$  , [\textbf{FAQs}]$_\checkmark$ , and a [\textbf{forum}]$_\checkmark$. \\
    E2: The [\textbf{faculty} in]$_\times$ OH was great and so was the administration . \\
    E3: [\textbf{ETrade}]$_\checkmark$ gives you a \$75 \textbf{bonus} upon establishing an account . \\
    \hline
    \end{tabular}
    \caption{Three different error types that BERT-Base+FMIM still has on the $\mathbb{L}\to\mathbb{S}$ setting. The words in bold are the ground truth aspect terms.}
    \label{tab: error}
\end{table*}

\section{Analysis}
\begin{figure}[t!]
    \centering
    \includegraphics[width=0.44\textwidth]{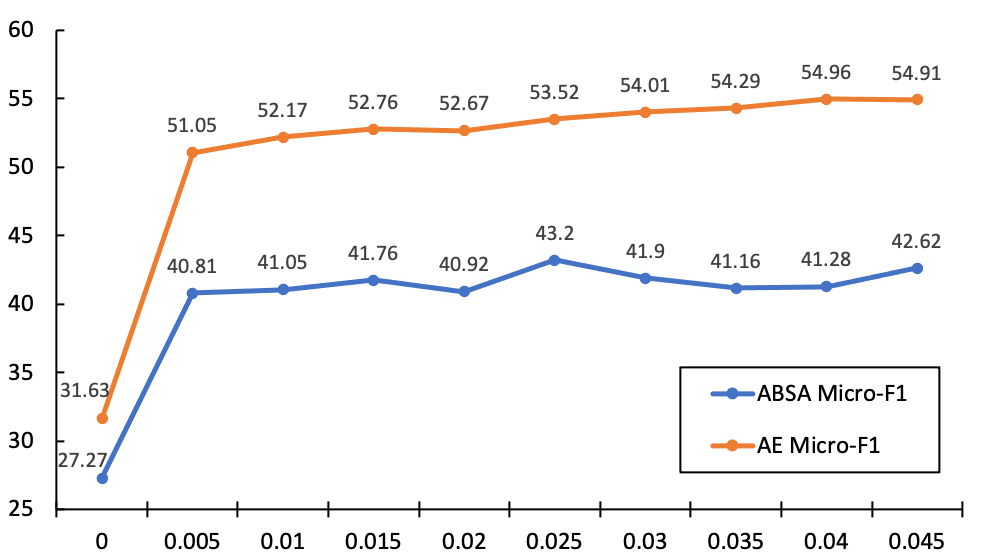}
    \caption{Results for ABSA and ATE with different $\alpha$.}
    \label{fig1}
\end{figure}

\begin{figure}[t!]
    \centering
    \includegraphics[width=0.44\textwidth]{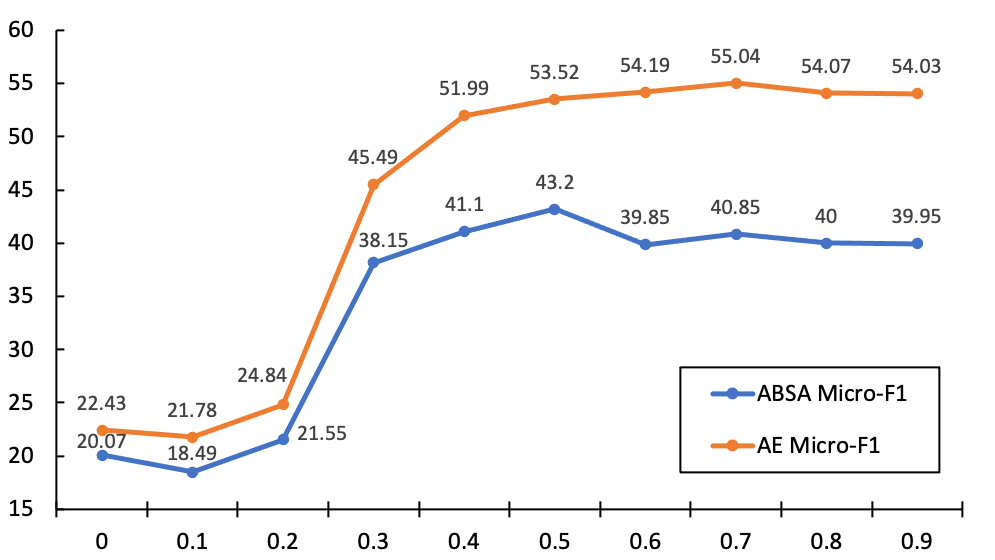}
    \caption{Results for ABSA and ATE with different $\rho$.}
    \label{fig2}
\end{figure}

\paragraph{Ablation Study} Since FMIM has only one single component, we can still investigate the effect of the hyperparameters. There are two crucial hyperparameters in our MI loss (see Eq~\ref{mi loss} and Eq~\ref{train loss}): the loss balancing factor $\alpha$ and the entropy threshold $\rho$. 
We test the performance of the model with different values of $\alpha$ and $\rho$ on $\mathbb{L}\to\mathbb{S}$  setting.

On one hand, we keep $\rho=0.5$ and alter the value of $\alpha$ in the range from 0 to 0.045. As illustrated in Figure~\ref{fig1}, our method degenerates into BERT-Base baseline when setting $\alpha=0$, which results in worse performance. For ABSA, the Micro-F1 reaches the peak (43.20) when setting $\alpha=0.025$. Moreover, we find FMIM's robustness to the change of $\alpha$. The performance keeps in a relatively stable range of 40.81$\sim$43.20 when varying $\alpha$ from 0.005 to 0.045. For ATE sub-task, we can observe that the performance maintains an upward trend with the increasing of $\alpha$. This demonstrates that FMIM's effectiveness in ATE task. However, since ATE is a sub-task, improving ATE does not necessarily improve ABSA. One can try to find a trade-off between them by tuning $\alpha$ carefully.

On the other hand, we keep $\alpha=0.01$ and change the value of $\rho$ from 0 to 0.9. As illustrated in Figure~\ref{fig2}, FMIM achieves the best performance when setting $\rho=0.5$. Similar to the phenomenon shown in Figure~\ref{fig1}, the performance of our model maintains stable when $\rho\ge 0.3$. FMIM collapses when $\rho\le0.2$, because setting an extremely small $\rho$ is equivalent to only optimizing the conditional entropy $H(\widehat{Y}|X)$ without optimizing $H(\widehat{Y})$, which may make the wrong predictions more confident. In practice, simply setting $\rho=0.5$ can observe a fairly competitive performance.

\paragraph{Case Study} In this section, we further study some cases to sufficiently illustrate our model's effectiveness qualitatively. With comparison to BERT-Base baseline, we calculate the two terms of mutual information (i.e. entropy of predicted labels $H(\widehat{Y})$ and conditional entropy $H(\widehat{Y}|X)$) to demonstrate the necessity of maximizing it.

Table~\ref{tab: case study} shows two sentences extracted from the service domain test data in $\mathbb{L}\to\mathbb{S}$ setting. For BERT-Base method, the model fails to give any predictions with a lower $H(\widehat{Y})$, a higher $H(\widehat{Y}|X)$ and a lower mutual information. While our FMIM method substantially increases the mutual information of the two sentences by 0.63 and 0.59, which lowers $H(\widehat{Y})$ and increases $H(\widehat{Y}|X)$. This leads to the correct final predictions. 

\paragraph{Error Analysis} We further study the errors that our approach still makes to provide some suggestions for future research. As shown in Table~\ref{tab: error}, there are three main error types. (1) discontinuous extraction, which may predict "glossary" and "terms" as aspects but omit "of" in the middle. (2) over-extraction, which may view the following "in" as part of the aspects. (3) under-recall, which may omit some aspects that require complex semantic relations. The inconsistent annotation of the dataset may also be a reason for this phenomenon.

\section{Conclusion}
In this paper, we propose using the fine-grained mutual information maximization (FMIM) technique to improve unsupervised domain adaptation for ABSA, ATE and NER. Our method is simple but has incredibly significant improvements over the strong baselines. 

The question of how to efficiently transfer domain knowledge remains unanswered. In the future, we plan to evaluate our method on more different tasks. Moreover, our proposed FMIM technique only introduces an additional loss term, which is orthogonal to all the previous domain adaptation methods for ABSA and NER. The effect of jointly using them still needs to be further explored.

\section*{Acknowledgments}
This work was supported by National Natural Science Foundation of China (61772036), Beijing Academy of Artificial Intelligence (BAAI) and Key Laboratory of Science, Technology and Standard in Press Industry (Key Laboratory of Intelligent Press Media Technology). Xiaojun Wan is the corresponding author. Thank the anonymous reviewers for the constructive suggestions. Thank Xinyu for useful discussions. Thank Yifan for paper revision.

\bibliography{custom}
\bibliographystyle{acl_natbib}



\end{document}